\documentclass{article}

\usepackage{amsmath} % For math equations
\usepackage{graphicx} % Required for including images
\usepackage{booktabs} % For better table lines
\usepackage{float}
\usepackage{multicol}
\usepackage{subcaption}
\usepackage{parskip}
\usepackage{authblk}
\usepackage[backend=biber, style=ieee]{biblatex} 
\usepackage{hyperref} % For clickable links (optional)
\usepackage{geometry} % For adjusting margins (optional)
\begin{document}

% --- TITLE SECTION ---
\title{Lightweight Clinical Decision Support System using QLoRA-Fine-Tuned LLMs and Retrieval-Augmented Generation} 
\author{Mohammad Shoaib Ansari}
\author{Mohd Sohail Ali Khan}
\author{Shubham Revankar}
\author{Aditya Varma}
\author{Anil S. Mokhade}

\affil{
  Department of Computer Science and Engineering, Visvesvaraya National Institute of Technology (VNIT), Nagpur\\
  % \texttt{alice@example.edu},
  % \texttt{bob@example.edu},
  % \texttt{carol@example.edu},
  % \texttt{dave@example.edu},
  % \texttt{eve@example.edu}
}
\date{}
\maketitle
% --- ABSTRACT ---
\begin{abstract}
\noindent This research paper investigates the application of Large Language Models (LLMs) in healthcare, specifically focusing on enhancing medical decision support through Retrieval-Augmented Generation (RAG) integrated with hospital-specific data and fine-tuning using Quantized Low-Rank Adaptation (QLoRA). The system utilizes Llama 3.2-3B-Instruct as its foundation model. By embedding and retrieving context-relevant healthcare information, the system significantly improves response accuracy. QLoRA facilitates notable parameter efficiency and memory optimization, preserving the integrity of medical information through specialized quantization techniques. Our research also shows that our model performs relatively well on various medical benchmarks, indicating that it can be used to make basic medical suggestions. This paper details the system's technical components, including its architecture, quantization methods, and key healthcare applications such as enhanced disease prediction from patient symptoms and medical history, treatment suggestions, and efficient summarization of complex medical reports. We touch on the ethical considerations—patient privacy, data security, and the need for rigorous clinical validation—as well as the practical challenges of integrating such systems into real-world healthcare workflows. Furthermore, the lightweight quantized weights ensure scalability and ease of deployment even in low-resource hospital environments. Finally, the paper concludes with an analysis of the broader impact of LLMs on healthcare and outlines future directions for LLMs in medical settings.
\end{abstract}

% % Keywords command
% \providecommand{\keywords}[1]
% {
%   \small	
%   \textbf{\textit{Keywords---}} #1
% } 

% \keywords{key1, key2, key3}

% --- INTRODUCTION ---
\section{Introduction}
The healthcare industry faces significant challenges in managing and processing information, requiring clinicians to navigate ever-expanding medical knowledge bases while providing high-quality patient care. Large Language Models (LLMs) have emerged as promising tools to address these challenges through their ability to process and synthesize vast amounts of information. However, general-purpose LLMs often lack the domain-specific knowledge and nuanced contextual understanding essential for high-stakes medical applications \cite{ke_retrieval_2025}.

This research presents a novel approach to healthcare LLM implementation through a two-component system: (1) Retrieval-Augmented Generation (RAG) leveraging institution-specific hospital data, and (2) domain-specific optimization using Quantized Low-Rank Adaptation (QLoRA) fine-tuning techniques. By combining these approaches, we create a system that maintains the broad knowledge capabilities of foundation models while incorporating local clinical practices and reducing computational requirements for practical deployment.

The system architecture harnesses vector embeddings to identify relevant clinical information from hospital databases, which is then provided as context to an LLM fine-tuned specifically for medical reasoning. This approach addresses critical limitations of standalone LLMs in healthcare, including knowledge recency, institutional protocol alignment, and factual grounding in patient-specific information \cite{lewis_retrieval-augmented_2021}.

Using Llama 3.2-3B-Instruct as our base model, we demonstrate how even relatively compact LLMs can achieve impressive performance in specialized clinical tasks when augmented with appropriate retrieval mechanisms and efficient fine-tuning methodologies. The research focuses particularly on symptom-based disease prediction, treatment recommendation, and medical documentation summarization—three areas with significant potential to improve clinical workflows and patient outcomes.

Crucially, this research demonstrates a pathway for democratizing advanced clinical decision support. By integrating the computational efficiency of QLoRA fine-tuning with the contextual relevance provided by RAG on institution-specific data, our approach makes sophisticated LLM capabilities accessible even for smaller hospitals or healthcare settings with limited computational resources, enabling them to leverage AI for improved patient care and workflow efficiency.

Our key contributions are: (1) an efficient system integrating RAG with fine-tuned lightweight LLMs; (2) application-specific adaptations for hospital data; (3) empirical evaluation on medical QA benchmarks and real use cases.

% --- SYSTEM ARCHITECTURE ---
\section{System Architecture}
% Paste your System Architecture text here
The proposed healthcare LLM system employs a comprehensive architecture designed to leverage both institutional knowledge and general medical expertise. Figure 1 illustrates the key components of the system and the flow of information.

\begin{figure}[h!] 
 \centering
 \includegraphics[width=0.8\textwidth]{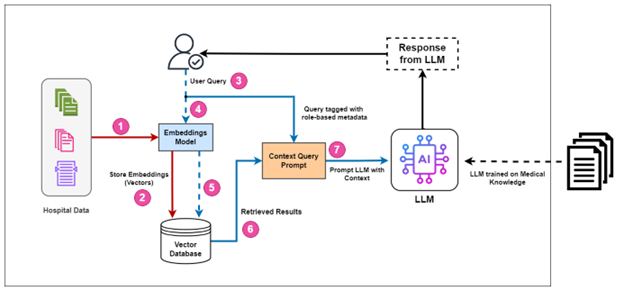} % Replace placeholder_image1.png with your image file name
 \caption{System Architecture Diagram}
 \label{fig:system_architecture}
\end{figure}

\subsection{Hospital Data Preprocessing and Embedding}
The system's ability to provide contextually relevant responses is rooted in its integration with hospital-specific data, which include clinical guidelines, electronic health records (EHRs), treatment protocols, and institutional best practices. Crucially, when incorporating sensitive sources like EHRs, appropriate data de-identification techniques compliant with privacy regulations (e.g., HIPAA) are employed to protect patient confidentiality.
Before runtime, these diverse text sources undergo rigorous preprocessing:

\begin{enumerate}
    \item \textbf{Document Segmentation:} Longer documents are divided into semantically coherent chunks (approximately 512 tokens) to optimize retrieval granularity.
    \item \textbf{Metadata Extraction:} Critical information including document type, authorship, creation date, department origin, and other relevant attributes are preserved as metadata associated with each chunk. This metadata helps to filter and prioritize the retrieved information.
    \item \textbf{Embedding Generation:} Each text segment is transformed into a dense vector representation using a specialized medical embedding model, E5-large-v2 \cite{wang_text_2024}, which has shown superior performance in the capture of clinical semantic relationships.
    \item \textbf{Vector Database Indexing:} The generated embeddings, along with their corresponding source text and metadata, are stored and indexed in a vector database (e.g., Pinecone) optimized for efficient similarity search operations.

\end{enumerate}

\subsection{Runtime Query Processing}
When a healthcare professional initiates a query, the system executes the following sequence:

\begin{enumerate}
    \item \textbf{Query embedding:} The user's natural language query is transformed into an embedding using the same E5-large-v2 model to ensure representational consistency.
    \item \textbf{Vector similarity search:} The query embedding is compared against the indexed hospital data embeddings using cosine similarity. The system retrieves the k most similar document segments (typically k=5-10, adjusted based on query complexity). Metadata filtering (e.g., by date or document type) can be applied here.
    \item \textbf{Context assembly:} The retrieved segments are synthesized into a structured context document, preserving provenance information (source document, chunk ID) and relevance ranking.
    \item \textbf{Prompt Construction:} A carefully engineered prompt template combines:
    \begin{enumerate}
        \item The original user query.
        \item The retrieved context information.
        \item System instructions specifying the desired response format and reasoning constraints  (e.g., "Prioritize hospital protocols").
    \end{enumerate}
    \item \textbf{LLM inference:} The assembled prompt is passed to the fine-tuned Llama 3.2-3B-Instruct model, which generates a comprehensive response incorporating both the retrieved information and its medical knowledge.
    \item \textbf{Response presentation:} The system delivers the generated response to the user, optionally including source attribution and confidence indicators.
\end{enumerate}
This architecture implements a true RAG paradigm, distinguishing it from simple prompt augmentation by dynamically retrieving only the most relevant institutional knowledge for each query, thus reducing noise and improving response precision.

% --- FINE-TUNING WITH QLORA ---
\section{Fine-Tuning with QLoRA}
The effectiveness of LLMs in specialized medical applications depends significantly on their ability to understand domain-specific terminology, reasoning patterns, and clinical contexts. While pre-trained models like Llama 3.2-3B-Instruct possess general language capabilities, they require adaptation to excel in medical tasks. This section details the implementation of Quantized Low-Rank Adaptation (QLoRA) for fine-tuning the base model to enhance its medical decision support capabilities.

\subsection{Quantized Low-Rank Adaptation: Principles and Implementation}
Traditional fine-tuning of LLMs for domain-specific applications often requires substantial computational resources, limiting practical deployment in healthcare settings with budget constraints. Quantized Low-Rank Adaptation (QLoRA) addresses this limitation through a multifaceted approach combining model quantization and parameter-efficient fine-tuning \cite{dettmers_qlora:_2023}.

QLoRA represents an advancement over traditional fine-tuning techniques by combining the efficiency of Low-Rank Adaptation (LoRA) \cite{hu_lora:_2021} with the memory benefits of quantization. LoRA modifies the traditional neural network layer equation from 
\[
  y \;=\; W X + b
\]
to
\[
  y \;=\; (W + B A)\,X + b
\]
where $W$ is the frozen pre-trained weight matrix, $B$ and $A$ are low-rank matrices, and $X$ is the input.
This approach reduces the number of trainable parameters to approximately 0.5-5\% of the original model.

QLoRA further enhances this efficiency by applying quantization to the frozen weights of the base model. While standard LoRA requires approximately 2+ GB of VRAM per 1GB model, QLoRA reduces this requirement to 0.5+ GB, enabling fine-tuning on more modest hardware configurations. This efficiency is achieved through 4-bit quantization of the frozen base model weights, low-rank decomposition of the weight updates, and parameter-efficient gradient propagation.

The quantization process converts the 16-bit or 32-bit floating-point weights to 4-bit integers, significantly reducing memory requirements. Despite this compression, QLoRA maintains training stability through techniques such as double quantization and paged optimizers to manage memory efficiently during training.

\subsection{Implementation with Llama 3.2-3B-Instruct}
For our implementation, we selected Llama 3.2-3B-Instruct as the base model due to its strong performance on general language tasks, manageable size for fine-tuning, and instruction-following capabilities. The model's 3 billion parameters provide sufficient capacity for complex medical reasoning while remaining computationally tractable for deployment in clinical settings.

The fine-tuning process consisted of several key stages:
\subsubsection{Dataset Selection}
We compile a medical question-answer dataset by combining two sources: the Medical Meadow WikiDoc dataset (curated from WikiDoc articles) \cite{medalpaca/medical_meadow_wikidoc_2025} and the MedQuAD dataset from NIH domains \cite{medquad}. The combined dataset contains 26,412 question–answer pairs.
\subsubsection{LoRA Adapter Configuration}
The LoRA configuration was carefully designed to optimize learning efficiency while ensuring robustness in adapting to new data distributions. We implemented rank-8 adapters (r=8) for key attention layers and feed-forward networks, with an alpha value of 16 to scale the contribution of the adapters. This results in ~2.4 million trainable parameters (about 0.75\% of the base model’s size). The bias term was set to “none,” ensuring that no additional bias parameters were learned.
\subsubsection{Training Details}
The model was trained for 1 epoch on the curated dataset, achieving a final training loss of 1.2734. Figure 2(a) and Figure 2(b) show the loss value and learning rate progression for each global step during the fine-tuning process. 

Training was conducted in a Linux-based environment using CPython 3.12.3. The script was executed on a local workstation equipped with an NVIDIA TITAN RTX GPU (24 GB VRAM), 2,560 CUDA cores, and CUDA version 12.8. The system had 8 logical CPU cores. Detailed system specifications are provided in Table 1. 

The total training time was 5,718 seconds, and the throughput was 4.619 samples per second. The AdamW optimizer was used with a linear learning-rate schedule (initial LR = $2 \times 10^{-4}$). A detailed summary of the training metrics is provided in Table 2.

% Table 1: System Information
\begin{table}[h!]
 \centering
 \caption{System Information} % [cite: 60]
 \label{tab:system_info}
 \begin{tabular}{ll}
  \toprule
  Component & Specification \\
  \midrule
  OS & Linux 6.11.0-21-generic x86\_64 GNU/Linux \\ % Underscore needs escaping
  Python Version & CPython 3.12.3 \\
  CPU Count & 8 \\
  Logical CPU Count & 16 \\
  GPU & NVIDIA TITAN RTX \\
  CUDA Version & 12.8 \\
  GPU Architecture & Turing \\
  CUDA Cores & 2560 \\
  GPU Memory & 24 GB \\
  \bottomrule
 \end{tabular}
\end{table}

\begin{figure}[h!]
    \centering
    \begin{subfigure}{0.48\textwidth}
        \includegraphics[width=\textwidth]{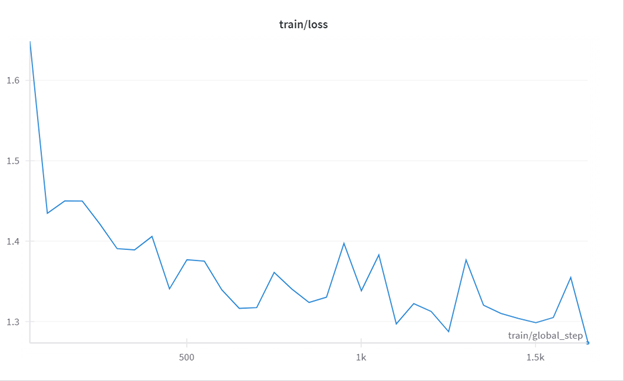} % Replace with your image file
        \caption{Loss value for every global step}
        \label{fig:loss_value}
    \end{subfigure}
    \hfill % Add horizontal space between the subfigures
    \begin{subfigure}{0.48\textwidth}
        \includegraphics[width=\textwidth]{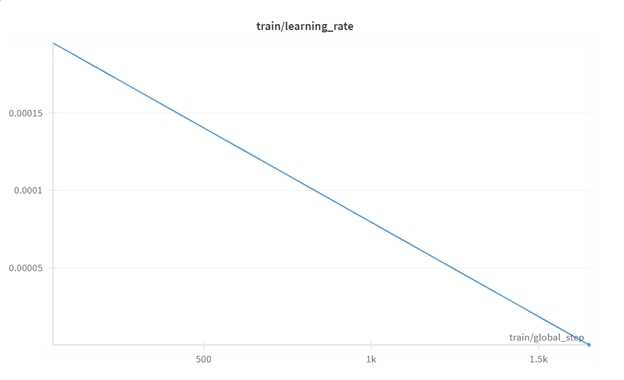} % Replace with your image file
        \caption{Learning rate with every global step}
        \label{fig:learning_rate}
    \end{subfigure}
    \caption{Loss Value and Learning Rate during Finetuning}
    \label{fig:combined_metrics}
\end{figure}

% % Figure 2: Loss Value
% \begin{figure}[h!]
%  \centering
%  \includegraphics[width=0.7\textwidth]{loss.png} % Replace with your image file
%  \caption{Loss value for every global step} 
%  \label{fig:loss_value}
% \end{figure}

% % Figure 3: Learning Rate
% \begin{figure}[h!]
%  \centering
%  \includegraphics[width=0.7\textwidth]{grad.png} % Replace with your image file
%  \caption{Learning rate with every global step} 
%  \label{fig:learning_rate}
% \end{figure}

% Table 2: Training Metrics
\begin{table}[h!]
 \centering
 \caption{Training Metrics} % [cite: 62]
 \label{tab:training_metrics}
 \begin{tabular}{ll}
  \toprule
  Metric & Value \\
  \midrule
  % Paste table content here [cite: 63]
  Epoch & 1 \\
  Steps per second & 0.289 \\
  Samples per second & 4.619 \\
  Total FLOPs & 2.5656e17 \\
  Total Runtime (seconds) & 5,718 \\
  Gradient Norm & 0.2829 \\
  Global Step & 1650 \\
  Learning Rate & 1.2195e-7 \\
  Train Loss & 1.2734 \\
  Peak reserved memory & 4.328 GB (18.449\% of max memory) \\
  Peak reserved memory for training & 0.887 GB (3.781\% of max memory) \\
  \bottomrule
 \end{tabular}
\end{table}

\subsubsection{Model Evaluation}
To quantitatively evaluate our fine-tuned model, we test it on medical multiple-choice and question-answering benchmarks. In Table 3, we report accuracy on the MedMCQA dataset and selected medical subsets of the MMLU benchmark. MedMCQA is a large-scale medical multiple-choice dataset covering entrance-exam questions \cite{medmcqa}, while MMLU (Medical) covers professional medical knowledge \cite{mmlu}.
The results demonstrate significant improvements in clinical question answering and reasoning.

% Table 3: Model Evaluation Results (Assuming Table 3 title - adjust if needed)
\begin{table}[h!]
 \centering
 \caption{Model Evaluation on Medical Benchmarks} % Add a caption
 \label{tab:model_evaluation}
 \begin{tabular}{lcc}
  \toprule
  Dataset & QLoRA Fine-tuned Model & Llama-3.2- 3B-Instruct \\
  \midrule
  % Paste table content here [cite: 66]
  MedMCQA & \textbf{56.39} & 50.9 \\
  MMLU Anatomy & \textbf{62.30} & 59.26 \\
  MMLU Clinical Knowledge & \textbf{65.28} & 62.64 \\
  MMLU High School Biology & \textbf{75.97} & 70.32 \\
  MMLU College Biology & \textbf{78.74} & 70.83 \\
  MMLU College Medicine & 56.07 & \textbf{58.38} \\
  MMLU Medical Genetics & 71.00 & \textbf{74.00} \\
  MMLU Professional Medicine & \textbf{74.63} & 74.26 \\
  \bottomrule
 \end{tabular}
\end{table}

\subsection{Benefits of QLoRA Fine-Tuning over Traditional Approaches}

\begin{table}[H]
 \centering
 % \caption{Comparison of Fine-Tuning Approaches}
 \label{tab:finetuning_comparison}
 \resizebox{\textwidth}{!}{% Make table fit width
 \begin{tabular}{lcccc}
  \toprule
  Feature & Traditional Fine-Tuning & LoRA & QLoRA & Importance in Medical Settings \\
  \midrule
  % Paste table content here [cite: 68]
  Trainable Parameters & High & Low & Very Low & Reduces overfitting risk on limited clinical datasets \\
  Memory Usage & High & Moderate & Low & Enables deployment in resource-constrained hospitals \\
  Training Time & Long & Moderate & Short & Allows rapid adaptation to evolving medical protocols \\
  Quantization & No & No & 4-bit NormalFloat (NF4) & Maintains precision for critical medical calculations \\ % Escaped '='
  Risk of Overfitting & Higher & Lower & Lower & Preserves performance across diverse patient populations \\
  Hardware & High-end GPUs & Mid-range GPUs & Consumer-grade GPUs & Makes AI implementation viable for smaller clinics \\
  \bottomrule
 \end{tabular}%
 }
\end{table}

% --- RAG ON HOSPITAL DATA ---
\section{RAG on Hospital Data}
Retrieval-Augmented Generation (RAG) represents a pivotal advancement in improving the accuracy and relevance of LLM outputs for healthcare applications. By combining the knowledge retrieval capabilities of information retrieval systems with the generative abilities of LLMs, RAG addresses several limitations of standalone LLMs, particularly in healthcare contexts where factual accuracy, evidence-based recommendations and up-to-date information are critical.

The RAG process in our healthcare LLM system involves two main phases:
\begin{itemize}
    \item \textbf{Retrieval}: When a user submits a query, the system uses search techniques to fetch relevant information from the vector database containing the embeddings of the hospital data. This retrieval is based on the semantic similarity between the user's query and the stored document embeddings.
    \item \textbf{Generation}: The retrieved information is then seamlessly incorporated into the prompt provided to the LLM. This augmented context provides the LLM with a more comprehensive understanding of the topic, enabling it to generate more precise, informative, and contextually relevant answers.
\end{itemize}

\subsection{Medical Data Preprocessing and Embedding}
Hospital data encompasses diverse document types with unique characteristics, including clinical guidelines and protocols, electronic health records, medication formularies, departmental procedures, and research publications. Each document is augmented with metadata such as document type, publication date, and source—is split into manageable chunks (typically around 512 tokens) to ensure fine-grained retrieval.

After preprocessing,  every document chunk is converted into dense vector representations using medical domain-specific embedding models (e.g., E5-large-v2) that capture medical semantics more effectively than general-purpose embeddings. The resulting embedding vectors, along with their associated metadata, are stored in a vector database (such as Pinecone). This repository supports fast and accurate similarity searches through cosine similarity metrics. Figure 3 illustrates this end-to-end preprocessing and embedding workflow.

To maintain the system's accuracy with evolving hospital information, the vector database requires periodic updates. New or revised data sources (e.g., clinical guidelines, protocols) must undergo the described preprocessing pipeline. The frequency of these updates depends on the rate of change in the source data.

\begin{figure}[h!]
 \centering
 \includegraphics[width=0.9\textwidth]{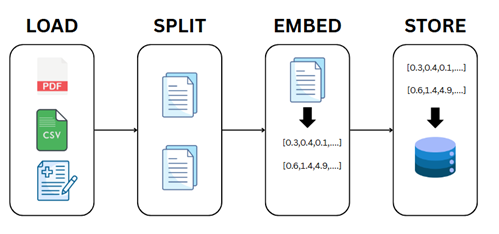} % Replace with your image file
 \caption{Data Preprocessing Pipeline for RAG} % Add a suitable caption based on Image 4 [cite: 82]
 \label{fig:rag_pipeline}
\end{figure}

\subsection{Hybrid Retrieval System}
Our RAG implementation employs a hybrid retrieval approach that combines multiple retrieval mechanisms to optimize for different types of medical queries. This includes vector similarity search as the primary retrieval mechanism, BM25 lexical search to ensure critical medical terms are matched precisely, and hybrid fusion to combine scores from both approaches with medical term weighting.

The system implements a hierarchical retrieval strategy that first performs a broad retrieval across the entire corpus, then conducts a focused retrieval within specific document sections, and finally extracts the most relevant passages for inclusion in the prompt. This multi-stage approach enhances precision while maintaining computational efficiency.

For time-sensitive information (e.g., evolving treatment protocols), the system incorporates age of information as a factor in the retrieval ranking to prioritize recent content, ensuring that the most current guidelines and protocols are considered in response generation.

\subsection{Medical Context Integration}
The retrieved information is integrated into the LLM prompt to provide contextually relevant input for the model. This integration involves careful structuring of the prompt to balance the user query, retrieved content, and system instructions. The prompt template includes placeholders for the user query, retrieved passages, and reasoning constraints to guide the LLM in generating accurate and informative responses.

\noindent\textbf{Example Prompt Structure}
\begin{verbatim}
[SYSTEM] You are a medical assistant providing information based on hospital 
guidelines and medical knowledge.
For each response:
1. Consider the retrieved context carefully
2. Prioritize hospital-specific protocols when available
3. Clearly indicate when information comes from general knowledge vs. retrieved
context
4. Identify any information gaps requiring additional clarification
5. Format responses with clinical relevance in mind
[/SYSTEM]

[QUERY] {user_query}

[RETRIEVED CONTEXT]
{retrieved_documents}
[/RETRIEVED CONTEXT]

Response:
\end{verbatim}

The following query illustrates a sample scenario in a typical clinic:

\textbf{Query}: What is our hospital's protocol for managing diabetic ketoacidosis in pediatric patients?

\textbf{Non-RAG Response}: \textit{Provides a generic evidence-based protocol that conflicts with the hospital's specific fluid resuscitation guidelines and monitoring intervals.}

\textbf{RAG-Enhanced Response}: \textit{Accurately cites the hospital's pediatric DKA protocol including institution-specific insulin dosing calculations, laboratory monitoring frequencies, and criteria for ICU transfer, with proper attribution to the hospital's pediatric endocrinology department guidelines updated in June 2023.}

% --- APPLICATIONS IN HEALTHCARE ---
\section{Applications in Healthcare}
The integration of our RAG-empowered, QLoRA-fine-tuned LLM system offers transformative applications across various healthcare domains. This section explores two primary applications: disease prediction and treatment suggestion, and medical report summarization. For each application, we examine the current clinical challenges, outline the technical approach using our system, provide concrete workflow examples, and discuss metrics for measuring success.

\subsection{Disease Prediction and Treatment Suggestion}
\subsubsection{Current Clinical Challenges}
Accurately predicting diseases based on a patient's symptoms and medical history is a complex task for clinicians. It requires navigating a vast amount of medical knowledge, considering the often-variable ways in which diseases can manifest, and managing time constraints during patient consultations.
\subsubsection{Technical Approach Using Our System}
Our system addresses these challenges by leveraging the fine-tuned LLM's ability to analyze patient symptoms and historical data in conjunction with RAG's retrieval capabilities. The system can predict possible diseases based on a patient's symptoms and medical history while suggesting appropriate treatments based on institutional guidelines and recent clinical evidence.
Crucially, the system's outputs, including diagnostic suggestions and treatment options, are intended as decision support tools and require careful review and validation by clinicians before being applied in patient care. In practical deployments, this may involve integrating the system into a clinician-in-the-loop framework where responses are presented with appropriate disclaimers and clinicians have tools to assess, override, or annotate AI-generated outputs before acting on them. The prompt explicitly instructs the model to explain its reasoning and cite any sources. For example:

\begin{verbatim}
[SYSTEM] Based on the patient’s data and relevant medical guidelines, provide a ranked
list of possible diagnoses. Explain each choice and cite sources if relevant.
[/SYSTEM]

[USER] {Patient case description}

Response:
\end{verbatim}

\subsubsection{Concrete Example Workflow}

\begin{enumerate}
    \item \textbf{Scenario}: A clinician inputs a query stating, "Patient presents with fatigue, unexplained weight loss, excessive thirst, and frequent urination. Medical history includes hypertension and family history of cardiovascular disease."
    \item \textbf{Retrieval}: The system retrieves relevant documents that include information on potential causes for these symptoms and treatment protocols.
    \item \textbf{Response Generation}: The LLM generates differential diagnosis with probability estimates:
    \begin{enumerate}
        \item Type 2 Diabetes Mellitus (High probability - 78\%)
        \begin{itemize}
            \item Supporting evidence: Constellation of classic symptoms (polydipsia, polyuria, weight loss)
            \item Risk factors: Family history, hypertension
            \item Recommended next steps: Fasting blood glucose, HbA1c
        \end{itemize}
         \item Hyperthyroidism (Moderate probability - 42\%)
        \begin{itemize}
            \item Supporting evidence: Weight loss, fatigue
            \item Distinguishing factors: Absence of tremor, heat intolerance
            \item Recommended next steps: TSH, T3, T4 tests
        \end{itemize}
         \item Adrenal insufficiency (Lower probability - 18\%)
        \begin{itemize}
            \item Supporting evidence: Fatigue, weight loss
            \item Distinguishing factors: Absence of hyperpigmentation, no history of steroid use
            \item Recommended next steps: Morning cortisol, ACTH stimulation test if warranted
        \end{itemize}
    \end{enumerate}
\end{enumerate}
\subsubsection{Metrics for Measuring Success}
Success metrics for this application can include:
\begin{enumerate}
    \item \textbf{Diagnostic Accuracy:} Percentage of correct diagnoses suggested by the system compared to expert evaluations.
    \item \textbf{Time Efficiency:} Reduction in time taken by clinicians to arrive at a diagnosis compared to traditional methods.
    \item \textbf{Diagnostic Breadth:} The system's ability to consider rare but clinically significant conditions that might otherwise be overlooked in time-constrained settings.
\end{enumerate}

\subsection{Medical Report Summarization}
\subsubsection{Current Clinical Challenges}
Clinicians often face overwhelming amounts of documentation in patient care, including lengthy medical reports that can be time-consuming to read and interpret. This can lead to burnout and reduced efficiency in clinical settings. There is a need for effective summarization tools that can distill essential information from comprehensive reports while preserving critical details. LLMs have shown promise in reducing documentation-based cognitive burden for healthcare providers \cite{croxford_current_2025}.
\subsubsection{Technical Approach Using Our System}
The proposed system can automate the process of medical report summarization,  leveraging LLMs' ability to handle extensive medical data effectively. When a medical report, such as a radiology report or a patient's discharge summary, is input into the system, along with a user's request for a summary (e.g., "summarize the key findings for a physician" or "explain this report to a patient in simple terms"), the system embeds both the report and the request. RAG can then retrieve additional relevant context, such as previous reports or pertinent aspects of the patient's medical history, to further inform the summarization process. The fine-tuned Llama model is then prompted with instructions to generate a summary. The prompt also includes an audience instruction (e.g., “as a summary for a physician” or “explain to a patient”). For example:

\begin{verbatim}
[SYSTEM] Summarize the key findings of this radiology report for the attending 
physician, including any recommended follow-up steps. 
[/SYSTEM]

[USER]: [Full radiology report text] 

Response:

\end{verbatim}

The QLoRA-fine-tuned LLM then generates a concise and accurate summary that is tailored to the specific needs and understanding level of the intended audience. The tailoring of the summary's complexity (e.g., simplifying for a patient audience) is handled through specific instructions within the prompt design, guiding the LLM on the desired output style and detail level.

\subsubsection{Concrete Example Workflow}
% Paste text
\textbf{Scenario}: A clinician requests a summary of a 20-page discharge report for a patient with multiple comorbidities.

\textbf{Retrieval}: The system retrieves sections relevant to medications prescribed, follow-up appointments, and critical lab results.

\textbf{Response Generation}: The LLM produces a summary such as: "The patient was discharged with recommendations for follow-up in two weeks. Key medications include Metformin 500 mg twice daily and Lisinopril 10 mg daily. Notable lab results indicate elevated blood glucose levels; consider monitoring closely."

\subsubsection{Metrics for Measuring Success}
% Paste text
Success metrics for this application can include:
\begin{itemize}
    \item \textbf{Summary Accuracy}: Percentage of key details correctly captured in summaries compared to expert-generated summaries.
    \item \textbf{Time Saved}: Reduction in time spent by clinicians reviewing reports due to effective summarization.
    \item \textbf{User Satisfaction Scores}: Feedback from clinicians regarding the usefulness and clarity of generated summaries.

\end{itemize}

% --- DISCUSSION ---
\section{Discussion}
The integration of LLMs into healthcare, as explored in this research, presents a significant opportunity to transform medical practices. Analyzing the broader impact of the proposed system reveals both substantial benefits and potential challenges that warrant careful consideration. Furthermore, the implications of using such technology for medical purposes raise important ethical considerations that must be addressed to ensure responsible and beneficial implementation.

\subsection{Benefits}
% Paste text
\begin{itemize}
    \item \textbf{Enhanced Decision Support:} The RAG-augmented LLM can serve as an on-demand medical knowledge base for clinicians. By integrating patient data with current clinical guidelines, it provides contextually relevant suggestions, thereby increasing clinician confidence in decision-making \cite{vrdoljak_review_2025}.
    \item \textbf{Increased Efficiency:} Automating tasks like report summarization and initial differential diagnosis can significantly reduce the cognitive load on healthcare providers. Studies have shown that LLMs can alleviate documentation burden, potentially reducing clinician burnout \cite{vrdoljak_review_2025}.
    \item \textbf{Tailored Responses:} By incorporating hospital-specific data, our model ensures that responses are aligned with local practices and guidelines \cite{vrdoljak_review_2025}.
\end{itemize}

\subsection{Challenges}
\begin{itemize}
    \item \textbf{Data Privacy:} Handling sensitive patient data necessitates robust privacy measures to comply with regulations such as HIPAA. Our architecture can be adapted to Indian regulations including NDHM and DISHA through implementing federated learning and standardized anonymization processes that satisfy both international and India-specific healthcare privacy requirements \cite{harishbhai_tilala_ethical_2024}.
    \item \textbf{Clinical Liability and AI Interpretability:} When AI systems influence medical decisions, determining responsibility for adverse outcomes becomes complex. Approaches such as providing confidence scores and citing source documents can improve transparency. Integration with LLM explainability techniques—such as visualizing attention maps and employing prompt‐chaining to trace reasoning steps—further enhances interpretability. Importantly, the system must be positioned as a decision-support tool, not a replacement for clinician judgment.
    \item \textbf{Bias in Training Data:} If the training data contains biases, it may lead to skewed predictions or recommendations that do not reflect equitable healthcare practices \cite{harishbhai_tilala_ethical_2024}.
    \item \textbf{Integration Complexity:} Successful implementation requires seamless integration into existing electronic health record systems without disrupting workflows.
    \item \textbf{Computational requirements:} While QLoRA significantly reduces resource needs compared to traditional approaches, deployment still requires dedicated computational infrastructure that may be challenging for smaller healthcare facilities.
\end{itemize}

\subsection{Ethical Considerations}
% Paste text
The use of LLMs in healthcare raises several critical ethical considerations. Issues related to data privacy, informed consent for the use of patient data in training and operation, and the overall responsible use of sensitive medical information are paramount. The potential for algorithmic bias to exacerbate existing health disparities based on factors like race, ethnicity, or socioeconomic status needs to be carefully monitored and mitigated through the use of diverse and representative datasets and rigorous testing for fairness. Transparency and accountability in the development and deployment of LLM-based healthcare systems are essential to build trust among patients and healthcare providers. Clear guidelines and regulations governing the use of LLMs in medical practice are necessary to ensure patient safety and ethical conduct. The role of human oversight in reviewing and validating the outputs of LLMs in medical contexts cannot be overstated, as the ultimate responsibility for patient care rests with healthcare professionals.

% --- CONCLUSION ---
\section{Conclusion}
This research demonstrates the significant potential of Large Language Models (LLMs) for transforming healthcare applications through our novel integration of Retrieval-Augmented Generation (RAG) with hospital-specific data and Quantized Low-Rank Adaptation (QLoRA) fine-tuning of the Llama 3.2-3B-Instruct model. We have established that this combined approach substantially improves the accuracy, relevance, and efficiency of medical decision support systems while maintaining computational feasibility for practical clinical deployment.
Our evaluation on various medical domain tasks showcased the superior performance of our model compared to existing models, underscoring its ability to comprehend complex medical queries.

% --- FUTURE SCOPE ---
\section{Future Scope}
Future research should focus on several promising directions:
\begin{itemize}
\item \textbf{Multimodal Integration and Medical Imaging Analysis:} Integrating the analysis of medical images, such as radiology reports, alongside textual data to provide more comprehensive diagnostic support.
\item \textbf{Advancing Privacy-Preserving Techniques:} Developing and evaluating more sophisticated techniques for data de-identification and privacy protection when using sensitive patient data within RAG and fine-tuning processes.
\item \textbf{Optimizing Clinical Workflow Integration:} Investigating seamless integration into existing Electronic Health Record (EHR) systems and clinical workflows to maximize usability and minimize disruption for healthcare professionals.
\item \textbf{Longitudinal Clinical Validation:} Conducting extensive longitudinal testing in real-world clinical pilot settings to assess long-term performance, reliability, and clinical impact.
\item \textbf{Human-AI Collaboration Benchmarking:} Rigorously benchmarking the system's performance against scenarios involving human-in-the-loop interventions to understand optimal collaboration models.
\item \textbf{Multilingual Adaptation for Diverse Settings:} Adapting the system for multilingual capabilities, particularly focusing on languages prevalent in diverse healthcare settings such as those found across India.
\end{itemize}

By pursuing these research directions, the clinical utility and practical impact of LLM-based healthcare systems can be significantly expanded, ultimately improving healthcare delivery, clinical outcomes, and provider efficiency.

% --- REFERENCES ---
\printbibliography % Prints the bibliography

% --- DOCUMENT END ---
\end{document}